\documentclass[journal]{IEEEtran}
\usepackage{amsmath,amsfonts}
\usepackage{algorithmic}
\usepackage{algorithm}
\usepackage{array}
\usepackage[caption=false,font=normalsize,labelfont=sf,textfont=sf]{subfig}
\usepackage{textcomp}
\usepackage{stfloats}
\usepackage{url}
\usepackage{verbatim}
\usepackage{graphicx}
\usepackage{cite}
\usepackage{hyperref}
\usepackage{booktabs}
\usepackage{color}
\usepackage{placeins}

\begin{document}

\title{Stereo-based 3D Anomaly Object Detection for Autonomous Driving: A New Dataset and Baseline}

\author{Shiyi Mu, Zichong Gu, Hanqi Lyu, Yilin Gao and Shugong Xu,~\IEEEmembership{Fellow,~IEEE}

\thanks{Manuscript received August 19, 2025; revised xx xx, 2025. This work was supported in part by  the National High Quality Program under Grant TC220H07D, in part by the National Key R\&D Program of China under Grant 2022YFB2902002, in part by the Innovation Program of Shanghai Municipal Science and Technology Commission under Grant 20511106603. The Associate Editor for this article was XXX. (Corresponding author: Shugong Xu.)}
\thanks{Shiyi Mu, Zichong Gu, Hanqi Lyu, Yilin Gao and Shugong Xu are with School of Communication \& Information Engineering at Shanghai University, Shanghai University, Shanghai 200444, China (e-mail: shiyimu@shu.edu.cn; guzichong123@shu.edu.cn; lvhanqi@shu.edu.cn; gaoyilin@shu.edu.cn; shugong@shu.edu.cn).}}

\markboth{Journal of \LaTeX\ Class Files,~Vol.~14, No.~8, August~2025}%
{Shell \MakeLowercase{\textit{Mu et al.}}: Stereo-based 3D Anomaly Object Detection for Autonomous Driving: A New Dataset and Baseline}


\maketitle

\begin{abstract}
3D detection technology is widely used in the field of autonomous driving, with its application scenarios gradually expanding from enclosed highways to open conventional roads. For rare anomaly categories that appear on the road, 3D detection models trained on closed sets often misdetect or fail to detect anomaly objects. To address this risk, it is necessary to enhance the generalization ability of 3D detection models for targets of arbitrary shapes and to possess the capability to filter out anomalies. The generalization of 3D detection is limited by two factors: the coupled training of 2D and 3D, and the insufficient diversity in the scale distribution of training samples. This paper proposes a Stereo-based 3D Anomaly object Detection (S3AD) algorithm, which decouples the training strategy of 3D and 2D to release the generalization ability for arbitrary 3D foreground detection, and proposes an anomaly scoring algorithm based on foreground confidence prediction, achieving target-level anomaly scoring. In order to further verify  and enhance the generalization of anomaly detection, we use a 3D rendering method to synthesize two augmented reality binocular stereo 3D detection datasets which named KITTI-AR. 
KITTI-AR extends upon KITTI by adding 97 new categories, totaling 6k pairs of stereo images. The KITTI-AR-ExD subset includes 39 common categories as extra training data to address the sparse sample distribution issue. Additionally, 58 rare categories form the KITTI-AR-OoD subset, which are not used in training to simulate zero-shot scenarios in real-world settings, solely for evaluating 3D anomaly detection.
Finally, the performance of the algorithm and the dataset is verified in the experiments. (Code and dataset can be obtained at https://github.com/shiyi-mu/S3AD-Code).
\end{abstract}


\section{Introduction}

\begin{figure}
\centering
\includegraphics[width=3.0in]{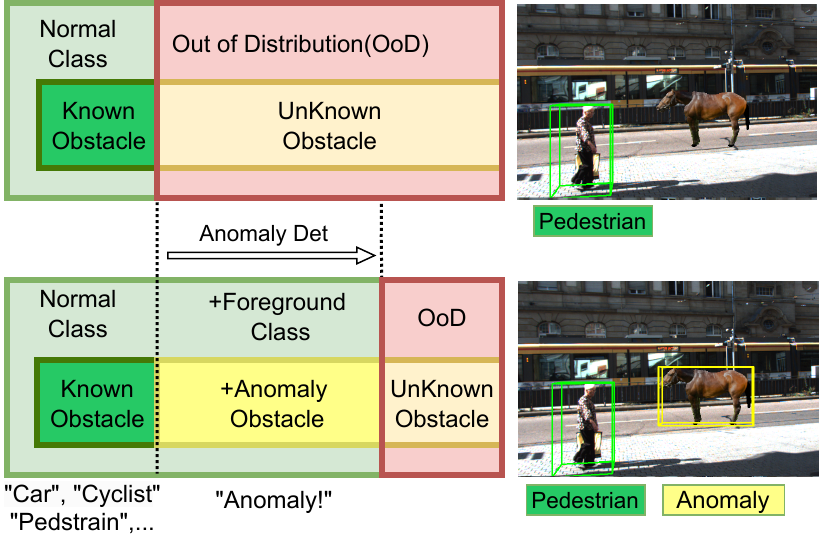}
\caption{3D Object Detection (Upper) and 3D Anomaly Detection (Lower). Green boxes indicate regular categories, and yellow boxes indicate anomaly categories.}
\label{fig_1}
\end{figure}

\IEEEPARstart{S}{afe} driving is very important in Intelligent Transportation Systems. Risks arising from the randomness of the environment and the limitations of algorithms fall within the scope of the Safety of The Intended Functionality (SOTIF). Environmental factors are determined by the Operational Design Domain (ODD), such as sensor performance degradation due to rain or snow, sudden changes in lighting when entering or exiting tunnels, and the need to use autonomous or assisted driving functions cautiously under foreseeable extreme weather conditions. Most existing road perception solutions are based on closed-set training. The limitations of algorithms are often manifested in their inability to effectively detect long-tail rare categories and Out-of-Distribution (OoD) new categories, which can result in critical failures such as the inability to trigger automatic emergency braking. For long-tail categories, performance can typically be improved by increasing the amount of training data collected or by adjusting the loss function. For unregistered new categories, anomaly detection or OoD detection methods are commonly used. As accidents caused by missed or incorrect detection occur, both academia and industry are increasingly focusing on research into road anomaly detection\cite{vojir2021road, ohgushi2020road, RBA_ICCV23, Mask2Anomaly_ICCV23, Mask2Anomaly_TPAMI24} and general obstacle detection. As shown in Figure~\ref{fig_1}, conventional 2D and 3D object detection algorithms can only detect the labeled known categories (green). The road anomaly detection algorithm detects more anomaly obstacles (yellow) from the foreground, reducing the area of Out-of-Distribution (red) to enhance safety.

Current road anomaly detection algorithms are developed based on two types of perception models: 2D anomaly bounding box detection\cite{peng2023sotif, heidecker2021towards} and 2D anomaly segmentation\cite{vojir2021road, ohgushi2020road, liu2024ensemble}. For object detectors, there are two possible issues with detecting anomalous objects: classes confusion and missed detection. High-risk objects might be confused as known low-risk categories, such as identifying a dynamic worker in an orange clothing is confused as a static traffic cone\cite{peng2023sotif}. This could lead the subsequent planning module to make more aggressive path choices, failing to leave enough safe clearance distance. Missed detection poses an even greater collision risk, such as mistaking a garbage bin in the middle of the road. For missed detection caused by a new category, class-agnostic foreground detectors can effectively improve this issue\cite{feng2022promptdet}. These detectors first identify every possible foreground object candidate boxes, which are denser and aim to cover both known and unknown objects as much as possible. Any object in the foreground that is not a normal object is considered an anomaly. The ultimate development of foreground detection leads to open-world detection\cite{cheng2024yolo-word} or open-vocabulary detection\cite{OV-Uni3DETR}, expanding the detector's category scale to locate any object in the open world.
For normal segmentation networks, each pixel is classified with a specific label. Similarly to detectors, if the pixels of an unknown object are classified as normal objects, it is a missed detection. Existing road anomaly segmentation methods predict anomaly objects at the pixel level. Typically, a panoptic segmentation model trained on existing road datasets\cite{RBA_ICCV23,Mask2Anomaly_ICCV23, Mask2Anomaly_TPAMI24} or image reconstruction algorithms\cite{Di_Biase_CVPR2021,ohgushi2020road,vojir2021road} perform uncertainty analysis on the output.

Corner cases in autonomous driving scenarios can be categorized into five levels\cite{breitenstein2021corner}: pixel-level, domain-level, object-level, scene-level, scenario-level. Current research on road anomaly detection mainly discusses two levels of anomalies: object-level and scene-level. 
Object-level anomalies are typically caused by unknown novel categories, such as animals and road obstacles. They can also consist of unknown configurations of known categories, such as overturned trucks. Scene-level anomalies are caused by abnormal spatial relationships or contexts involving known categories. For instance, trees in the middle of a road. Regardless of whether the corner cases are caused by normal or anomalous categories, the risk assessment is based on the presence of collision risks. Current road anomaly detection, which relies on 2D object detectors or 2D segmentation, can calculate the direction relationship between the anomalous object and the vehicle's direction of motion, but does not consider distance information. The lack of distance and obstacle scale information can lead to a failure to evade in a timely manner or result in unnecessary abrupt braking. As 3D detection and Bird's Eye View (BEV) detection technologies become more widely adopted, the implementation of 3D anomaly detection is deemed crucial. 3D road anomaly detection faces two major issues: dataset and low-cost detection algorithms.

To validate the 2D anomaly perception capability, the test dataset can be divided into two types: object detection based and segmentation based. For detection-based datasets, such as PeSOTIF\cite{peng2023pesotif}, a large number of monocular images containing anomalous scenes are collected and annotated with 2D bounding boxes. Segmentation-based data sets provide pixel-level segmentation annotations for anomalous objects, such as Lost-and-Found(LaF)\cite{LOF} and Road Anomaly\cite{RoadAnomaly}. Currently, there is a lack of large-scale annotated anomaly scenes in 3D anomaly detection datasets. Therefore, there are three ways to construct test datasets: category omission based, simulation based, and image editing based. The category omission method\cite{OV-Uni3DETR} involves discarding labels of categories like \textbf{\textit{Pedestrian}} from the KITTI dataset. This approach is limited by the original labeled categories, with the assumed anomaly classes being neither extensive nor rare enough. CARLA\cite{CARLA} can simulate multi-modal road scene datasets, allowing static assets to be placed in the driving area to create anomalous scenes. However, there is a significant discrepancy between the virtual background and real data, with domain differences in image style. Image editing involves projecting new 3D models into real road datasets background. In order to preserve the authentic background style of the dataset to the greatest extent, we propose a richly categorized augmented reality 3D anomaly detection dataset named KITTI-AR, which involves editing the original KITTI stereo dataset, rendering stereo images for new categories, and providing category and 3D ground truth labels. This dataset is designed to train and validate the proposed 3D anomaly detection algorithm.

At the level of algorithm design, unlike 2D anomaly detection, 3D detection necessitates distance prediction for anomalous foreground objects. An expensive but straightforward approach is to estimate distances from LiDAR point clouds\cite{ODN3D}, while a cost-effective but challenging method is to estimate from monocular cameras. Stereo solutions achieve a balance between cost and difficulty, which is why a binocular stereo approach is adopted as the foundational framework for the method in this paper. 

3D anomaly object detection poses two primary challenges: (1) the detection of novel object categories in 2D space, and (2) the accurate estimation of depth and scale. Traditional closed-set training leads to overfitting to known textures and poor generalization of 3D predictions to unseen instances. To address these issues, we propose two complementary strategies: decoupling and extra samples.

Decoupling is implemented in two forms: (1) the separation of binary foreground classification from $N$ classification over known categories, and (2) the decoupling of 2D and 3D supervision. For the first, we introduce a category-agnostic binary classifier based on disparity features to distinguish foreground objects from background, which facilitates the detection of generic obstacles regardless of semantic class. An anomaly scoring mechanism is then applied to separate OoD from known categories. For the second, we decouple 2D and 3D supervision to allow training with extra 2D annotations alone, thus reducing labeling cost while improving detection performance on OoD categories.

Building on the decoupling design, we identify a key limitation: the sparse scale distribution of known categories constrains the model's ability to estimate 3D properties for unseen OoD objects. This often causes predicted object sizes to be biased toward a limited set of known scale ratios. To alleviate this, we incorporate an extra training set KITTI-AR-ExD containing more diverse object scales, thereby improving the generalization capability of 3D scale estimation under open-set conditions.
Therefore, the key contributions of this paper are as follows:
\begin{enumerate}

\item{We release a 3D anomaly detection dataset KITTI-AR. It is designed to analyze the limitations of closed-set training algorithms, alleviate the sparsity of scale distribution, and validate the feasibility of OoD 3D object detection.}

\item{We propose a \textbf{S}tereo-based \textbf{3}D \textbf{A}nomaly \textbf{D}etection (S3AD) algorithm, capable of predicting both the 3D location and scale of road anomalies.}

\item{ The proposed method introduces a foreground background binary classification branch based on disparity features, along with an OoD scoring strategy, to detect a broader range of unknown obstacles.}

\item{Thanks to the proposed decoupling strategy, the potential of 3D anomaly detection can be effectively unlocked at low annotation cost by introducing only additional 2D foreground labels.}

\end{enumerate}
\section{RELATED WORK}
This section reviews related work on 3D object detection and road anomaly detection.
\subsection{3D Object Detection}

3D object detection \cite{liang2021survey} aims to identify and locate objects in 3D space, serving as the foundation for advanced perception in autonomous driving systems and garnering attention from recent studies \cite{mao20233d, qian20223d, song2024robustness}. Current methods for 3D object detection are typically categorized based on their input types, which include Camera-based \cite{shi2021geometry,yao2023occlusion,zhang2021objects,kim2022boosting,chen2023monocular,gao2022camrl,haq2022one,zhang2023monodetr,yang2024monopstr,chang2018pyramid,sun2020disp,gao2023real,YOLOStereo3D,wang2023towards,li2023bevdepth,li2022bevformer,huang2022bevdet4d,li2024fast}, LiDAR-based \cite{sun2024exploiting, liu2023anchorpoint, an2023rs, shan2023focal, chang2023svdnet, wang2023collaborative, zhao2021transformer3d}, and multimodal methods \cite{xie2024ppf,ahmed2022smart,he2022stereo,yin2023gal,zhang2023fs}. The latter two often achieves more accurate perception results, utilizing LiDAR to directly acquire robust 3D information of the scene or enhancing detection performance through sensor fusion. Many autonomous driving systems adopt these methods to significantly improve long-range detection capabilities. However, this high performance also comes with a high sensor cost, which to some extent hinders the deployment and application of these methods in practical systems.

To address the cost constraints, existing camera-based methods have discussed detection issues under visual-only conditions with different camera setups, including monocular\cite{shi2021geometry,yao2023occlusion,zhang2021objects,kim2022boosting,chen2023monocular,gao2022camrl,haq2022one,zhang2023monodetr,yang2024monopstr}, stereo binocular\cite{chang2018pyramid,sun2020disp,gao2023real,YOLOStereo3D}, and surround-view multi-camera systems\cite{wang2023towards,li2023bevdepth,li2022bevformer,huang2022bevdet4d,li2024fast}. Monocular methods have lowest cost but face the challenge of depth estimation. To solve this issue, target-awared methods are directly built on 2d object detection\cite{zhao2019object} but try to enhance 3d object detection by modeling geometric constraints\cite{shi2021geometry,yao2023occlusion}, reinforcing depth estimation\cite{zhang2021objects,kim2022boosting} and introducing auxiliary knowledge\cite{chen2023monocular,gao2022camrl,haq2022one}. 
Based on DETR\cite{DETR}, some depth-assisted methods introduce depth-guided transformers to uncover the implicit 3D information, such as MonoDETR\cite{zhang2023monodetr} and  MonoPSTR\cite{yang2024monopstr}.

Balancing sensor cost and performance in a stereo based approach for disparity estimation\cite{chang2018pyramid,sun2020disp}. For more efficiency, SAS3D\cite{gao2023real} proposes a strategy to perform different sampling density in outer and inner region while YOLOStereo3D\cite{YOLOStereo3D} enhances anchor-based detection with stereo features. Surround-view multi-camera systems have small overlap between cameras that prejudice disparity estimation, but they offer a wider field of view with multiple monocular estimations assembling in a single frame. Therefore, their research focus is on view transformation\cite{wang2023towards,li2023bevdepth} and spatial-temporal fusion\cite{li2022bevformer,huang2022bevdet4d}, which may cause huge computing consumption. Although there are some lightweight solutions like Fast-BEV\cite{li2024fast}, multi-camera methods still face significant resource burdens. 

\subsection{Road Anomaly Detection}
In autonomous driving scenarios, corner cases can be categorized into pixel-level, domain-level, object-level, scene-level, and scenario-level\cite{breitenstein2021corner}. Currently, mainstream research focuses on object-level anomalies. Breitenstein et al.\cite{breitenstein2021corner} proposed dividing anomaly detection in autonomous driving scenarios into five technical approaches: reconstruction, prediction, generative, feature extraction, and confidence scores.  Daniel et al.\cite{bogdoll2022anomaly} also published a survey based on this classification. Some new research works can also be included within these categories. 

\textbf{Reconstruction and Generative}. Reconstruction-based and generative methods follow a principle: the model cannot learn to reconstruct or generate anomalous images from normal training images. A reconstruction network can reconstruct the original image from normal input images, and anomalous regions will also be reconstructed as normal images. The reconstruction will obscure the anomalous regions, leading to differences in the reconstruction. These differences are considered anomalies. Such methods are also widely used in the field of industrial anomaly detection.
JSR-Net\cite{vojir2021road} combines the reconstruction differences with the street scene segmentation predicted by segmentation model to compute anomalous regions. Ohgushi et al.\cite{ohgushi2020road} propose a method that combines reconstruction differences with segmentation entropy loss. Di Baise et al.\cite{Di_Biase_CVPR2021} utilize reconstruction uncertainty to analyze anomalies. Lis et al.\cite{lis2023detecting} use a sliding window approach with local erasure reconstruction rather than global reconstruction.

\textbf{Feature Extraction}. Anomalous regions in the feature space are analyzed, with their feature distances being relatively far from the features of the training set. A typical method in industrial anomaly detection is PatchCore\cite{patch_core}, where normal samples are segmented into patches for feature extraction and recorded into a memory bank. During testing, the nearest distance between the patch to be tested and the normal features in the memory bank is calculated one by one. If the patch is anomalous, the distance will be larger. Similar anomaly distance calculations are used at the global image level in road anomaly detection.
DeepRoad\cite{DeepRoad} performs feature extraction and dimensionality reduction based on VGG and PCA. RPL\cite{Liu_RPL_ICCV23} introduces feature-level anomaly analysis within a segmentation framework, aiding the segmentation head in distinguishing out of distribution objects.

\textbf{Confidence}. Estimating anomalies based on the uncertainty of network output confidence. This type of anomaly analysis can be divided into multi-model and single-model approaches. Multi-model approaches include Monte Carlo Dropout(MCD)\cite{gal2016dropout} and deep ensemble methods, while single-model approaches analyze output logits. Bayesian SegNet\cite{kendall2015bayesian} proposes estimating the uncertainty of segmentation networks based on MCD. Multiple samples are taken from the trained segmentation network through dropout layers, and multiple classification results are predicted. The higher the variance of the results, the greater the uncertainty about that region, indicating an anomalous target. Peng et al.\cite{peng2021uncertainty} propose introducing MCD into YOLOv3 to evaluate the uncertainty of output classification and regression results. Heidecker et al.\cite{heidecker2021towards} introduced MCD on Mask-RCNN to achieve similar estimation goals.

Due to the need for multiple samples and inferences with MCD\cite{gal2016dropout} algorithms, such as 20 times\cite{peng2021uncertainty} or 100 times\cite{heidecker2021towards}, significant computational delays can occur, making it challenging to apply in real-time systems. Peng et al.\cite{peng2023sotif} propose a road anomaly detection algorithm based on deep ensemble methods. They train five object detection models with the same structure but different parameters. By matching multiple output results through intersection over union matching, the variance in predictions from multiple networks for the same target serves as the uncertainty estimation. The five detection models are run in parallel on different GPUs, which optimizes the time required for uncertainty prediction. 

The aforementioned multi-model MCD\cite{gal2016dropout} and deep ensemble methods predict the uncertainty of anomalous targets across multiple parameters or models. In contrast, single-model approaches estimate the anomaly score by only one inference. The simplest strategy is to analyze the maximum confidence. Maximum Softmax Probability (MSP)\cite{MSP} uses the negative value of the maximum softmax output as the anomaly score. Rejected by All (RbA)\cite{RBA_ICCV23} proposes treating the multi-class mask level as multiple binary classifications, with the anomaly score being the sum of the probabilities of anomalous features being rejected by all known class heads. M2A\cite{Mask2Anomaly_TPAMI24} introduces MSP\cite{MSP} into the mask classification layer of Mask2Former\cite{cheng2022masked} for mask-level anomaly analysis.

Earlier 3D open set detection methods were primarily based on point cloud approaches such as MLUC\cite{cen2021open} and OSIS\cite{OSIS2020}. The 3D Object Discovery Network(ODN3D)\cite{ODN3D} introduces the concept of guiding visual open set 3D detection based on point cloud open set detection, which is a semi-supervised pseudo-labeling approach. OV-Uni3DETR\cite{OV-Uni3DETR} proposes a unified multimodal open set 3D detection framework that integrates point cloud and visual image, as well as indoor and outdoor scenes. OV-Mono3D\cite{OV-Mono3D} introduces the first open-vocabulary 3D monocular detection framework, bridging open-vocabulary 2D detection with 3D perception. It holds the potential to enable 3D detection of arbitrary obstacles.

\section{METHODOLOGY}
\subsection{Stereo 3D Object Detection Algorithm}

\begin{figure}
\centering
\includegraphics[width=3.3in]{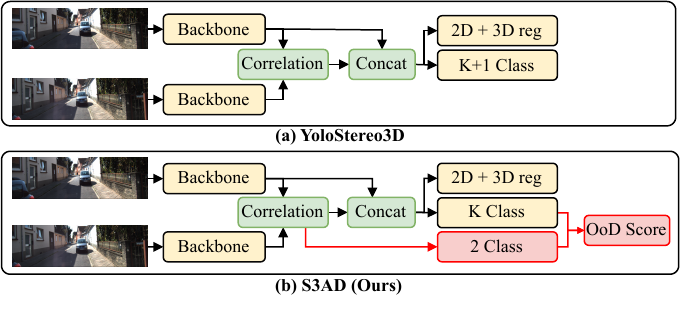}
\caption{Framework Comparison: (a) Stereo-based 3D object detection framework for closed-set, (b) Stereo-based 3D Anomaly Detection framework for open world.}
\label{fig_pipeline_compare_2}
\end{figure}

\begin{figure*}
\centering
\includegraphics[width=7.0in]{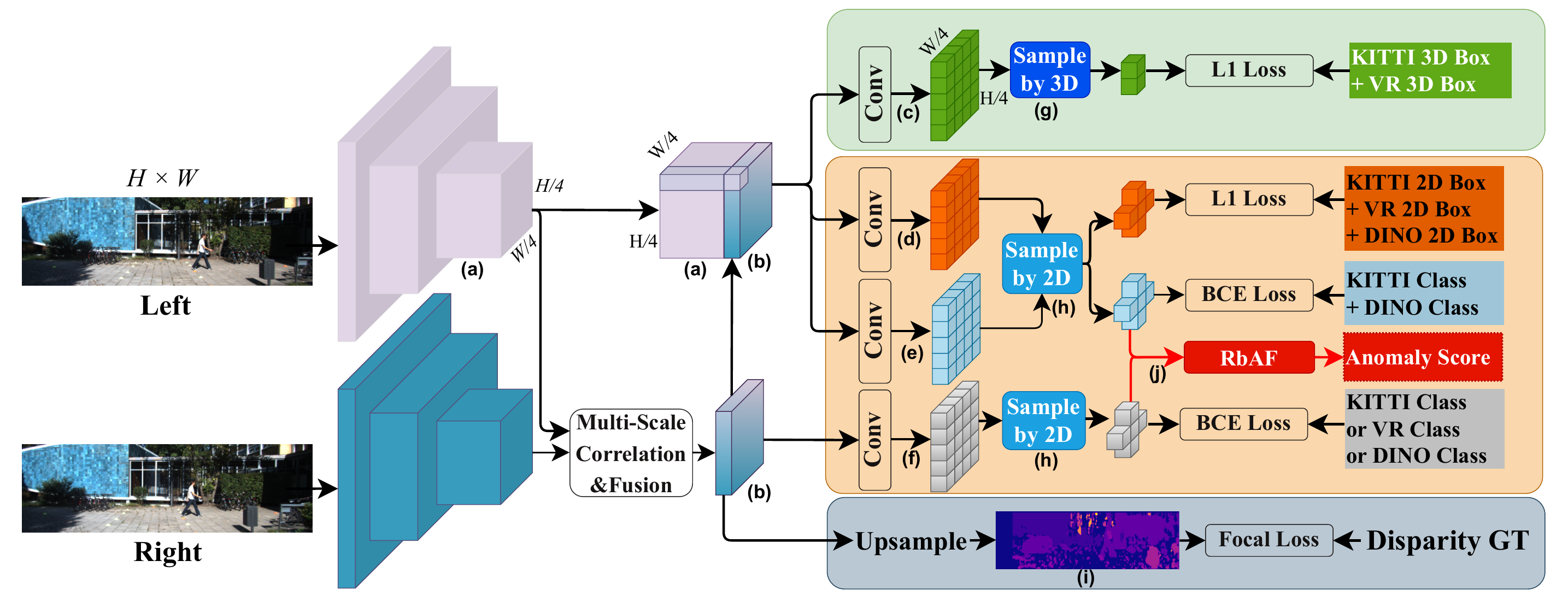}
\hfil
\caption{\textbf{Proposed method S3AD}. (a) Left view feature $f_L$, (b) Stereo features $f_s$, (c) 3D box regression head $H_{reg3D}$, (d) 2D box regression head $H_{reg2D}$, (e) normal category multi-classification head $H_{cls}$, (f) foreground binary classification head $H_{fg}$, (g) and (h) positive sampling based on 3D boxes and 2D boxes, (i) predict disparity $D$, (j) anomaly scoring.}
\label{fig_sim}
\end{figure*}

Based on the 3D detection framework YOLOStereo3D\cite{YOLOStereo3D}, we construct a stereo 3D anomaly detection framework. 
As shown in Figure~\ref{fig_pipeline_compare_2}, the main architectural difference between our method and YOLOStereo3D\cite{YOLOStereo3D} lies in the introduction of a new binary foreground classification branch based solely on disparity features.

The input stereo images are represented as a pair $(x_{L}, x_{R})$ where $x \in \mathbb{R}^{W\times H}$. The framework mainly consists of four parts: feature extraction backbone $F_b(\cdot)$, stereo multi-scale correlation fusion $F_s(\cdot)$\cite{YOLOStereo3D}, classification heads $H_{cls}(\cdot)$ and $H_{fg}(\cdot)$, regression heads $H_{reg2D}(\cdot)$ and $H_{reg3D}(\cdot)$, and disparity reconstruction $H_{dis}(\cdot)$. The feature of each single image is $f_L$ and $f_R$. 
The anchor-level prediction process of the network is as follows:
\begin{equation}
f_s = F_s\big(f_L, f_R\big),
\end{equation}
\begin{equation}
C_{norm} = H_{cls}\big([f_s, f_L]\big),
\end{equation}
\begin{equation}
Box_{2D} = H_{reg2D}\big([f_s, f_L]\big),
\end{equation}
\begin{equation}
Box_{3D} = H_{reg3D}\big([f_s, f_L]\big),
\end{equation}
where $f_s\in \mathbb{R}^{1152 \times W/4 \times H/4}$ is correlation fusion feature of stereo images. $ C_{norm}\in \mathbb{R}^{N \times K}$ is prediction of normal classes. $N$ is the number of regular categories and $K$ is the number of predefined anchors. $Box_{2D}\in \mathbb{R}^{4 \times K}$ is output of 2D box regression including $[x_{2d}, y_{2d}, w_{2d}, h_{2d}]$. $Box_{3D}\in \mathbb{R}^{8 \times K}$ is 3D box regression including 3D position $[x_{3d}, y_{3d} , z_{3d}]$, 3D scales $ [w_{3d}, h_{3d}, l_{3d}]$ and observation angle $[sin(2\alpha), cos(2\alpha)]$.
 
Unlike closed-set 3D object detection, OoD detection aims to localize novel categories or anomalies beyond known classes. We decompose this task into two sub-problems: (1) category-agnostic foreground detection and (2) anomaly scoring. The foreground detection is implemented as a binary classification head $H_{fg}(\cdot)$. Since disparity maps contain sufficient information to determine foreground regions, we exclusively feed stereo disparity features $f_s$ into this head to prevent overfitting to 2D appearance patterns, and calculate the foreground classification results of the anchor $C_{fg}\in \mathbb{R}^{1 \times K}$:

\begin{equation}
C_{fg} = H_{fg}\big(f_s \big).
\end{equation}

We design an OoD score function named RbAF (Rejected by All Foreground) that combines the foreground classification confidence $C_{fg}$ and multi-class confidence $C_{norm}$ of known categories.

\begin{equation}
C_{OoD} = \mathrm{RbAF}\big(C_{fg}, C_{norm} \big).
\end{equation} 

As an auxiliary task, an upsampling head $H_{dis}(\cdot)$ is used to predict the binocular disparity:
\begin{equation}
D = H_{dis}(f_s), D \in \mathbb{R}^{1 \times (W/4) \times (H/4)}.
\end{equation}

\subsection{Decoupled Supervision for Classification, 2D regression, and 3D regression} 

\begin{figure}
\centering
\includegraphics[width=3.2in]{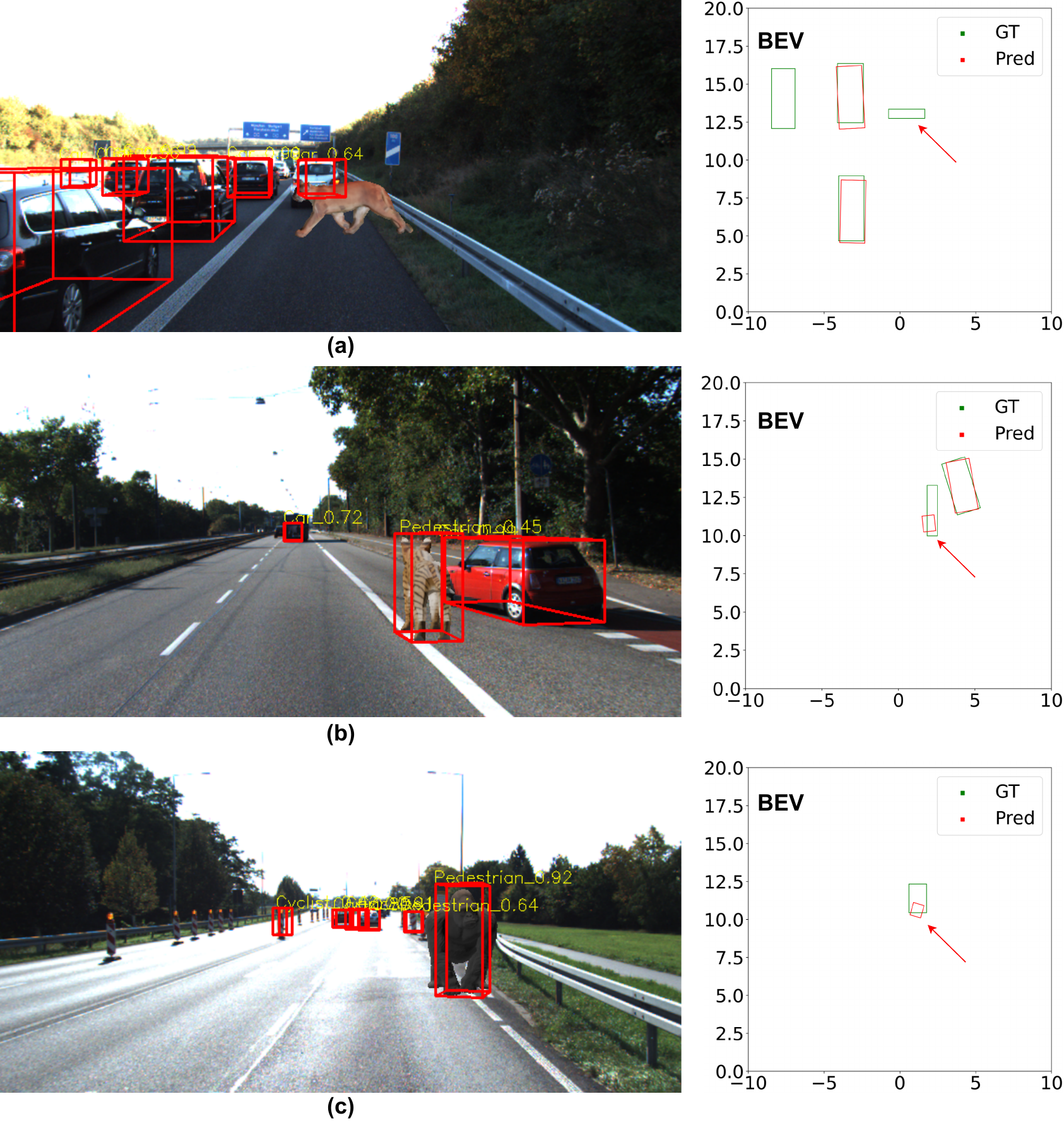}
\hfil

\caption{Visualization of the OoD test results of the model trained on KITTI normal categories only: (a) missed detection, (b) (c) misidentified as pedestrians and incorrectly estimated length and width scales.}
\label{fig_threshold}
\end{figure}

It is essential to analyze the performance of closed-set trained models under open-set conditions. We train the model on the original KITTI training set and perform inference on the KITTI-AR-OoD subset with a lowered detection confidence threshold. The visualization results are shown in Figure~\ref{fig_threshold}.
Clearly, for OoD categories, the model either misses detections or mistakenly classifies them as known categories with low confidence. Interestingly, as shown in the BEV (Bird's Eye View) visualization in Figure~\ref{fig_threshold} (b) and (c), although the scale estimation error for novel objects is notably large, their predicted positions are very close to the ground truth.

This observation suggests that the primary bottlenecks in generalization to OoD categories lie in the low confidence of foreground classification from 2D anchors and the inaccurate scale estimation. Enhancing 2D foreground detection capabilities and improving 3D scale estimation can significantly boost the model's generalization performance on OoD objects.
In addition to increasing real or synthetic 3D datasets to enhance the generalization of 2D foreground detection and 3D scale estimation, a more cost-effective alternative is to use easy 2D annotations or 2D open-world detection models\cite{OV-Mono3D}. To this end, we introduce a decoupled design that enables a supervision strategy under missing 3D annotations.

In detectors based on anchor such as Retina-Net\cite{retina} and YOLO series\cite{redmon2018yolov3}, anchors are sampled into positive and negative samples based on the Intersection over Union (IoU) relationship between the anchor-level output and the ground truth. The regression for both 2D and 3D is supervised by positive samples. Negative samples are labeled as the background for classification loss, but not for regression loss. Here, classification and regression are designed in a decoupled supervision scheme, meaning that the training features for the classification and regression heads are not consistent. The regression head is category-agnostic, thus maintaining the capability for 2D and 3D regression even for novel categories. 

In traditional sampling strategies, classification, 2D regression, and 3D regression share the same positive anchor assignment mechanism. To enable decoupling and leverage the extra 2D annotations, we adopt a dual sampling strategy. As shown in Figure~\ref{fig_sim}, the top-right area represents the 3D sampling process, while the middle-right area corresponds to the 2D sampling process. The number of annotated 3D and 2D boxes are denoted as $K'_{3D}$ and $K'_{2D}$, respectively. Since some annotations only contain 2D boxes, the number of 3D boxes can be less than that of 2D boxes. After applying the dual sampling strategy, the number of anchors involved in loss computation are denoted as $K_{3D}$ and $K_{2D}$, satisfying $K_{3D} \leq K_{2D}$. The 3D-sampled anchors are used for computing losses related to depth, orientation, and 3D scales, while the 2D-sampled anchors are used for 2D box regression, fine-grained classification, and binary foreground classification.

\subsection{Loss Functions}
The loss in the classification part consists of two components: Normal category multi-class classification loss $l^{norm}_{cls}$ and foreground binary classification loss $l^{fg}_{cls}$.

\begin{equation}
l^{norm}_{cls} = W_{norm} \cdot BCE(C_{norm}, C'_{norm}),
\end{equation}

\begin{equation}
W_{norm} = 
\begin{cases}
(1 - C_{norm})^{\gamma} & \text{if } C'_{norm} = 1, \\
(C_{norm})^{\gamma} & \text{otherwise},
\end{cases}
\end{equation}
where $C'_{norm} \in \mathbb{R}^{N \times (K_{pos} + K_{neg})} $ represents the ground truth for the normal categories, $ K_{pos} $  and $  K_{neg}$  represent the number of positive and negative anchors that have passed the IoU filtering. $W_{norm}$ represents the weights in the focal loss, $BCE(\cdot)$ is binary cross entropy. Similarly, the loss for foreground classification is a focal loss for a single binary classification:
\begin{equation}
l^{fg}_{cls} =  W_{fg} \cdot BCE(C_{fg}, C'_{fg}),
\end{equation}

Where $C'_{fg} \in \mathbb{R}^{1 \times (K_{pos} + K_{neg})} $ represents the foreground ground truth, determined by the original annotations and the pseudo-labels from the open vocabulary detector\cite{Groundingdino}. 

Positive and negative samples selected based on 2D IOU participate in the classification training, with only positive samples involved in the regression training. Since we have introduced 2D pseudo-labels for foreground annotations, some anchors lack 3D annotations. Therefore, two rounds of sampling are required: the first round samples anchors with valid 2D boxes as positive samples for 2D regression annotations $Box2D' \in \mathbb{R}^{4 \times K_{2D}}$, and the second round samples anchors with valid 3D boxes as positive samples for 3D regression training annotations $Box3D' \in \mathbb{R}^{8 \times K_{3D}}$. Where $K_{2D}$ and $K_{3D}$ represent the number of positive anchors for the two sampling rounds. It should be noted that $K_{3D} \leq K_{2D}$, because the 3D bounding boxes can be projected to a 2D boxes.
\begin{equation}
l^{2D}_{reg} = \frac{1}{K_{2D}}|Box_{2D} - Box_{2D}'|,
\end{equation}
\begin{equation}
l^{3D}_{reg} = \frac{1}{K_{3D}}|Box_{3D} - Box_{3D}'|.
\end{equation}

The supervision loss for disparity estimation follows the design of YOLOStereo3D\cite{YOLOStereo3D}, using stereo focal loss as the auxiliary supervision loss $l_{dis}$.
In summary, the total loss function for training is:
\begin{equation}
loss = \lambda_{1}l^{norm}_{cls} +  \lambda_{2}l^{fg}_{cls} +  \lambda_{3}l^{2D}_{reg} +  \lambda_{4}l^{3D}_{reg} + \lambda_{5}l_{dis}.
\end{equation}

The 3D scale annotations are normalized differently for each class in YOLOStereo3D\cite{YOLOStereo3D}, meaning that each class has its own mean and standard deviation. For outlier classes unknown to the training stage, it is impossible to determine which class mean and standard deviation to use for decode. Therefore, the training process in this paper only uses a unified mean and standard deviation for all known classes for normalization. During the anomaly category testing phase, decode with the same values.

\subsection{3D Anomaly Scoring for Objects}

For anchor-based detectors, the classifier predicts categories for each anchor. The dense anchor classification results are also filtered by Non-Maximum Suppression(NMS) to produce the final sparse results. Anchors with low confidence are filtered out. In 2D road anomaly segmentation methods\cite{Mask2Anomaly_TPAMI24, RBA_ICCV23, MSP}, anomaly scores can be obtained by analyzing the dense pixel-level confidence. A similar method can be designed based on the dense anchor-level outputs. 

A key concept is that unknown class objects in the foreground are considered anomalies. Based on the design of the foreground detector, we can obtain anchor-level foreground confidence score $c_{fg}\in \mathbb{R}^{1}$, normal class confidence score $c_{norm}\in \mathbb{R}^{N}$, and 2D and 3D bounding boxes. In 2D road anomaly segmentation, every pixel is assigned a clear class, even if it is road or sky. There are two methods by confidence analysis: based on the maximum confidence of known $N$ classes\cite{MSP} and based on the total confidence of all known classes\cite{RBA_ICCV23}. The formula for Maximum Softmax Probability (MSP) \cite{MSP} is defined as follows:
\begin{equation}
\operatorname{MSP}(x)=1-\max_{n=1}^{N}(softmax(c_{norm})).
\end{equation}

In the method based on the total confidence of all normal $N$ classes, Rejected by All (RbA)\cite{RBA_ICCV23}, the pixel-level anomaly score is the sum of the uncertainties for all normal class outputs. The formula for RbA is defined as follows:
\begin{equation}
\operatorname{RbA}(x)=1 - \frac{1}{N} \sum_{n=1}^{N} \sigma\left(c_{norm}\right),
\end{equation}
where $\sigma(\cdot) = \tanh(\cdot)$ in RbA\cite{RBA_ICCV23}. With the introduction of the foreground detection output, we define anchor-level anomaly score calculation method by Maximum Softmax Probability of Foreground (MSPF) and Rejected by All Foreground  (RbAF):

\begin{equation}
\operatorname{MSPF}(x)=c_{fg}-\max_{n=1}^{N}(softmax(c_{norm})),
\end{equation}
\begin{equation}
\operatorname{RbAF}(x)=c_{fg}-\frac{1}{N}\sum_{n=1}^{N} \sigma\left(c_{norm}\right),
\end{equation}
where $c_{fg}\in \mathbb{R}^{1}$ is anchor-level classification confidence score. $\sigma(\cdot) = sigmoid(\cdot)$ in our method. Since every pixel in 2D street panoramic segmentation has a clear label, All pixels can be classified as foreground classes, which is a special case where $c_{fg}=1$.

\subsection{3D Object Stereo Augmented Reality Dataset}

Currently, 2D road anomaly detection datasets are relatively well-developed, such as Lost-and-Found (LaF)\cite{LOF} and Road Anomaly\cite{RoadAnomaly}. However, 3D anomaly detection datasets are noticeably lacking. Compared to the potentially infinite number of categories in the open world, existing traffic scene 3D detection annotation categories are limited. For example, the KITTI\cite{KITTI} dataset includes only 8 categories. 3D detection evaluation metrics on the KITTI\cite{KITTI} dataset consider only three categories: \textit{\textbf{Car}}, \textit{\textbf{Pedestrian}}, and\textit{ \textbf{Cyclist}}. To fully train 3D object detectors or to determine open-world detection capabilities, a category-rich 3D detection dataset is needed. Researchers can simulate multi-modal autonomous driving perception datasets using the CARLA\cite{CARLA} and Unity\cite{virtual-KITTI}, but CARLA’s static and dynamic assets are still limited, and there are distribution differences between virtual and real scenes background. 

\begin{figure}
\centering
\includegraphics[width=3.0in]{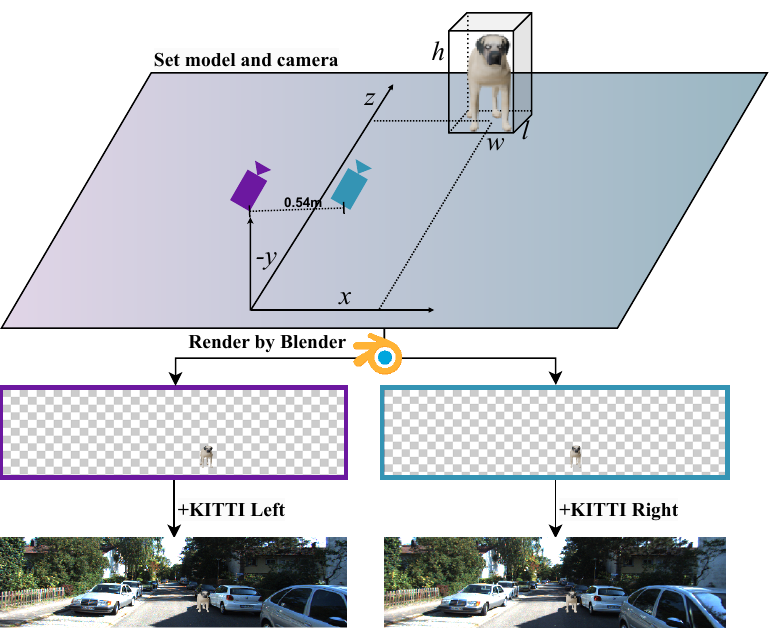}
\caption{The rendering process of the KITTI-AR new category dataset based on Blender.}
\label{blender}
\end{figure}

In 2D object detection tasks, dataset can be expanded by data collection or augmentation. For 3D object detection, annotating distance and 3D scale requires point clouds, which are costly to collect and annotate. Inspired by 2D detection’s use of cropping and pasting for data augmentation, 3D objects can be rendered onto background from existing datasets to extend them. The challenges in such rendering include setting virtual camera parameters and obtaining annotations for 3D objects. This paper constructs an augmented reality stereo 3D detection dataset named KITTI-AR. KITTI-AR is based on a large-scale 3D model dataset and the existing real-world 3D object detection dataset KITTI. The construction process is illustrated in Figure~\ref{blender}.

The construction process mainly consists of two parts: the setup of the Blender camera model and the acquisition of annotation. The positions, angles, FOVs, and resolutions of the left and right virtual cameras can be inferred from the original KITTI annotations. Based on the existing 3D objects in KITTI, the area where obstacles can be placed is calculated. The basic principle is that the placed obstacles should not be occluded by existing 3D objects in the 3D space. For newly placed 3D models, the scale $(w, h, l)$ and geometric center in the 3D space can be computed based on statistics for each mesh. These models are scaled to a reasonable size, translated to the target position and height, and converted to coordinates $(x, y, z)$ in the camera coordinate system. Finally, rendering is performed under random lighting conditions. The rendered stereo foreground images are overlaid with original KITTI stereo images to create the augmented reality dataset.

\begin{figure}[t]
\centering
\includegraphics[width=\columnwidth]{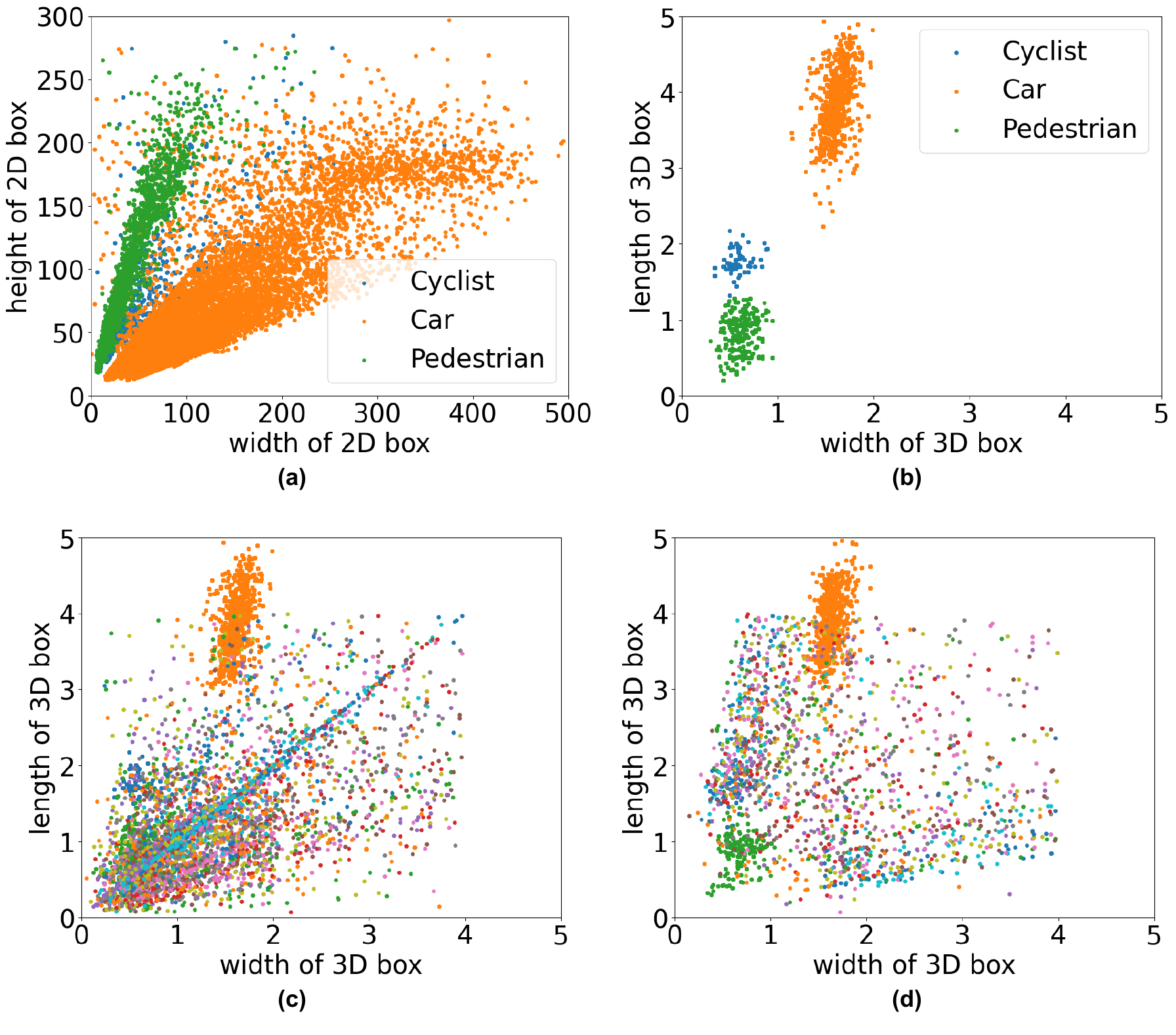}
\caption{
Scale distributions in the dataset: (a) width and height of 2D boxes in the original KITTI, (b) length and width of 3D boxes in the original KITTI, (c) length and width of 3D boxes in KITTI-AR-ExD, (d) length and width of 3D boxes in KITTI-AR-OoD.}
\label{size}
\end{figure}

For training and testing purposes, we have synthesized two subsets of KITTI-AR: the KITTI-AR-ExD training set and the KITTI-AR-OoD evaluation set. The background images for the training and evaluation sets come from the commonly used split method\cite{Chen_Split} in KITTI. The number of categories and the sample size are as shown in Table~\ref{Dataset Compare}. 
The KITTI-AR-ExD dataset utilizes 39 common indoor object categories, such as sofas, refrigerators, and printers. The KITTI-AR-OoD dataset employs other 58 categories of anomaly models, such as elephants, goats, fire hydrants, wheelchairs, and trash bins. To ensure a fair OoD evaluation, it is critical that the 58 categories in the KITTI-AR-OoD subset are used strictly for testing and are excluded from any training stage.

\begin{table}
\centering
\caption{Compare the categories and sample sizes of synthetic datasets with KITTI.}
\label{Dataset Compare}
\begin{tabular}{@{}c|ccc@{}}
\toprule
dataset                            & classes & train samples & val samples \\ \midrule
KITTI\cite{KITTI} & 8       & 3712          & 3769        \\
KITTI-AR-ExD      & 8+39      & 4038          & -           \\
KITTI-AR-OoD      & 8+58      & -             & 2347        \\ \bottomrule
\end{tabular}
\end{table}

The scale distribution of the original training samples in KITTI is relatively discrete and hardly covers the scales of the test categories, as shown in Figure~\ref{size}. The imbalance and lack of richness in the scale distribution of the training samples of KITTI led to the detection scale errors shown in Figure~\ref{fig_threshold} (c), where the length of new categories was predicted as the width and length of \textbf{\textit{Pedestrian}}.

\section{EXPERIMENTS}
This section presents the experimental setup, a performance comparison with existing methods on OoD data, qualitative visualizations, and ablation studies on key components of the proposed approach.

\subsection{Experimental Setup}
In terms of code framework and hyper-parameter selection, we align with YOLOStereo3D\cite{YOLOStereo3D}, choose PyTorch as the training and inference framework, with a single NVIDIA 4090D GPU with 24GB of memory. The optimizer is Adam with an initial learning rate of 0.0001. We employ a cosine learning rate schedule, and the training batch size is 8. The input images are cropped by removing the top 100 pixels in height and then resized to 288$\times$1280. The data augmentation settings are consistent with YOLOStereo3D\cite{YOLOStereo3D}. In all OoD tests, the categories used for OoD testing are completely excluded from the training process.

\subsection{Comparison with Existing Algorithms}

Table~\ref{kitti_single_ood_table} follows an evaluation design based on OV-Uni3DETR\cite{OV-Uni3DETR}, \textbf{\textit{Car}} and \textbf{\textit{Cyclist}} in KITTI are used for training, while the \textbf{\textit{Pedestrian}} is an OoD category that is excluded from training.
There are 3,712 training samples and 3,769 evaluation samples\cite{Chen_Split}. The report includes the $AP3D$ at 11 recall positions with 0.25 threshold on moderate difficulty subset\cite{OV-Uni3DETR}. In the table, \textbf{PC} denotes point cloud input, \textbf{Mono} indicates monocular camera input.

To the best of our knowledge, we are the first to propose a 3D anomaly detection algorithm for stereo vision. Most of algorithms listed in Table~\ref{kitti_single_ood_table} are all designed based on point clouds or monocular vision. Compared to the current state-of-the-art open-vocabulary 3D detection algorithms OV-Uni3DETR\cite{OV-Uni3DETR}, which combines point clouds with monocular vision, S3AD is able to achieve a comparable solution based on stereo vision. It should be noted that the cost of binocular sensors is much lower than LiDAR. To further compare the performance on monocular vision, we simply degrade the stereo solution by removing the feature extraction of the right view and replacing the feature disparity estimation module with a simple convolution layer. This simple degradation design can still surpass the open-set detection capability of the monocular OV-Uni3DETR\cite{OV-Uni3DETR}, even though the detection performance for normal categories has degraded to below the baseline. It should be noted that no synthetic datasets were used in our method in Table~\ref{kitti_single_ood_table}.

\begin{table}[hb]
\centering
\caption{Compared with existing open-set detection algorithms.
}
\label{kitti_single_ood_table}
\begin{tabular}{@{}c|c|ccc@{}}
\toprule
Method   & input   & OoD &Car&Cyc    \\ \midrule
Det-PointCLIP\cite{Det-PointCLIP}     & PC      & 0.32           & 3.67           & 1.32           \\
Det-PointCLIPv2\cite{Det-PointCLIPv2} & PC      & 0.32           & 3.58           & 1.22           \\
OV-Uni3DETR\cite{OV-Uni3DETR}         & PC      & \textbf{19.57}          & \textbf{92.4}4          & \textbf{56.67}          \\ \midrule
OV-Uni3DETR\cite{OV-Uni3DETR}         & Mono    & 9.98           & \textbf{75.14}          & \textbf{18.44}          \\
S3AD(Ours)                                   & Mono    & \textbf{15.36}          & 66.79          & 10.55          \\ \midrule
3D-CLIP\cite{CLIP}                    & PC+Mono & 1.28           & 42.28          & 21.99          \\
OV-Uni3DETR\cite{OV-Uni3DETR}         & PC+Mono & \textbf{23.04} & \textbf{92.55} & \textbf{58.21} \\ \midrule
S3AD(Ours)                                 & Stereo  & 21.37          & 80.93          & 23.09          \\ \bottomrule
\end{tabular}
\end{table}

\begin{table}
\centering
\caption{Test performance $AP_{3D}$ and $AP_{2D}$ on KITTI-AR-OoD, with the original KITTI data and KITTI-AR-ExD as training samples.}
\label{kitti_ar_animal_eval_2D_3D}
\begin{tabular}{@{}c|ccc@{}}
\toprule
                    & $AP_{OoD}$  & $AP_{ped}$    & $AP_{car}$    \\
training data       & $3D/2D$       & $3D/2D$       & $3D/2D$       \\ \midrule
KITTI-train only  & 9.09 / 9.09   & 38.48 / 49.76  & 79.28 / 81.70 \\
+KITTI-AR-ExD 2D & 21.05 / 87.89 & 43.95 / 61.15 & 79.96 / \textbf{88.88} \\
+KITTI-AR-ExD 3D & \textbf{74.35} / \textbf{90.06} & \textbf{48.26} / \textbf{64.80} & \textbf{80.20} / 88.53 \\ \bottomrule
\end{tabular}
\end{table}

Can the experiments in Table~\ref{kitti_single_ood_table} sufficiently demonstrate that our algorithm is ready for practical applications? There are limitations in performing anomaly detection testing on the original KITTI dataset. The reason is that the newly assumed \textbf{\textit{Pedestrian}} (anomaly) and the \textbf{\textit{Cyclist}} (normal) have similar textures and 3D scales. The detection capability for unknown objects of \textbf{\textit{Pedestrian}} may be derived from the spillover of \textbf{\textit{Cyclist}}. To better evaluate performance on a wider range of anomaly objects, we use KITTI-AR-OoD as the test dataset. \textbf{\textit{Pedestrian}}, \textbf{\textit{Cyclist}}, and \textbf{\textit{Car}} in KITTI serve as the basic normal training set. The test results can be seen in the first row of Table~\ref{kitti_ar_animal_eval_2D_3D}. 
The 3D detection AP on OoD drops to 9.09\%, which indicates poor performance under the 11-recall evaluation metric.
This phenomenon indicates that the anomaly detection evaluation method designed in Table~\ref{kitti_single_ood_table} cannot fully expose the shortcomings of the 3D OoD detection algorithm.

Thanks to the decoupling strategy designed in our framework, we are able to enhance the model's generalization to arbitrary foreground objects by introducing additional binary foreground annotations and class-agnostic 3D box supervision. As shown in Table~\ref{kitti_ar_animal_eval_2D_3D}, after incorporating only the 2D annotations from the KITTI-AR-ExD subset, the 3D detection AP on OoD increases to 21.00\%, and the 2D detection AP on OoD improves to 87.89\%, even though the novel categories in the test set never appear during training. With further inclusion of 3D bounding boxes from KITTI-AR-ExD, the 3D $AP_{OoD}$ on OoD rises significantly to 74.35\%. Meanwhile, the performance on regular in-distribution categories also benefits, with the \textbf{\textit{Pedestrian}}  class achieving a 9.78\% gain in AP\textsubscript{3D}.

To better evaluate the effectiveness of our proposed method, we re-implement the recent open-world 3D detection algorithm OV-Mono3D\cite{OV-Mono3D} on the KITTI-AR-OoD dataset. As shown in Table~\ref{kitti_ar_ood_compare-ov-mono3d}, OV-Mono3D\cite{OV-Mono3D} incorporates the open-vocabulary 2D detector\cite{Groundingdino}, achieving slightly better 2D detection performance on OoD compared to our approach. However, due to the limited generalization ability of monocular depth estimation, its 3D detection performance on OoD is significantly lower than that of our proposed S3AD.

\begin{table}
\centering
\caption{Test performance $AP_{3D}$ and $AP_{2D}$ on KITTI-AR-OoD.}
\label{kitti_ar_ood_compare-ov-mono3d}
\begin{tabular}{@{}c|c|ccc@{}}
\toprule
Method      & IoU          &  $AP2D_{OoD}$ & $AP3D_{OoD}$ \\ \midrule
OV-Mono3D\cite{OV-Mono3D}   & \textgreater 0.05  &\textbf{90.91}   & 5.39  \\
S3AD(Ours)  & \textgreater 0.05  &90.14   & \textbf{87.04} \\ \midrule
OV-Mono3D\cite{OV-Mono3D}  & \textgreater 0.25  &\textbf{90.91}   & 0.64\\
S3AD(Ours)  & \textgreater 0.25  &90.06   & \textbf{74.35} \\ \bottomrule 

\end{tabular}
\end{table}

\begin{figure*}[!ht]
\centering
\includegraphics[width=7.0in]{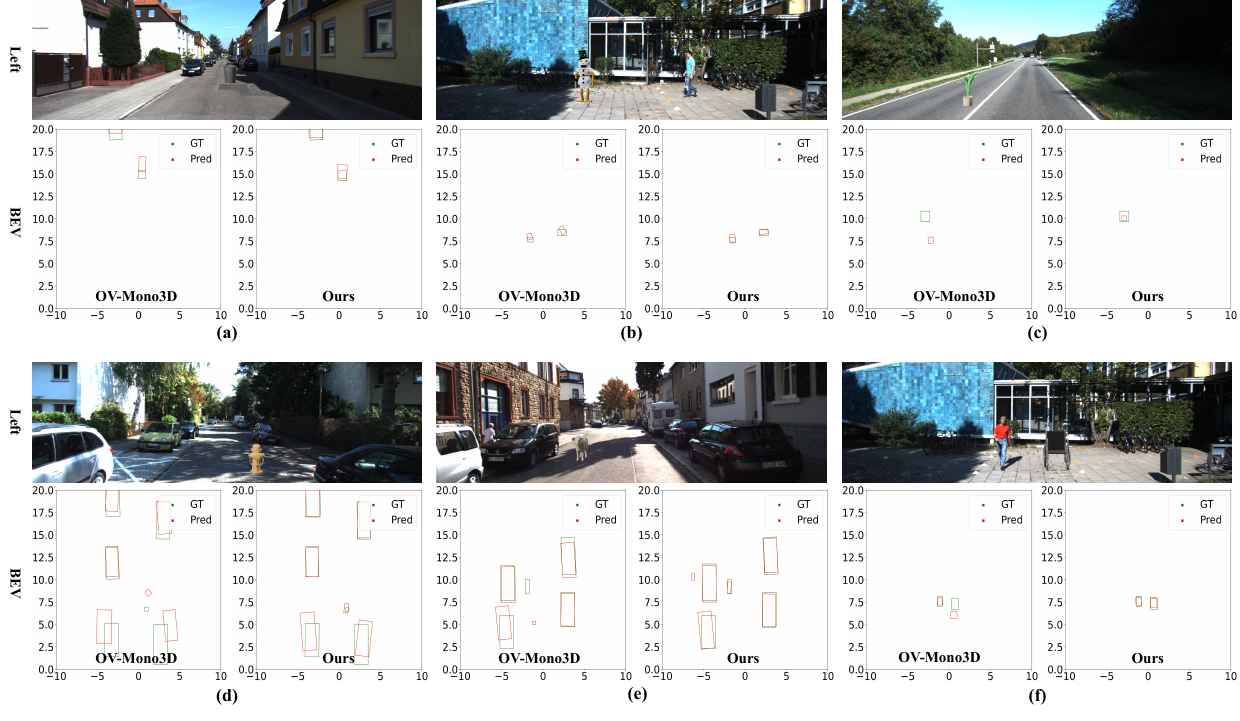}
\hfil
\caption{Comparison with OV-Mono3D on BEV results for OoD categories: (a) trash bin, (b) snowman, (c) flower pot, (d) fire hydrant, (e) sheep, and (f) wheelchair.}
\label{fig_results_bev}
\end{figure*}

To provide a more intuitive comparison between our method and OV-Mono3D\cite{OV-Mono3D} in terms of 3D estimation, we present BEV (Bird's Eye View) visualizations in Figure~\ref{fig_results_bev}. The subfigures illustrate representative OoD objects, including a trash bin, snowman, flower pot, fire hydrant, sheep, and wheelchair. It can be observed that while OV-Mono3D\cite{OV-Mono3D} achieves roughly correct orientation and size estimation for anomalous objects, our method demonstrates a significant advantage in distance (depth) estimation accuracy.

\subsection{Ablation Study}

\noindent \textit{1) Benefit from the decoupling of 2D and 3D.}

In Table~\ref{kitti_ar_animal_eval_2D_3D}, it can be observed that by simply adding more 2D annotations of irrelevant foreground classes, the performance of 3D detection for OoD (anomaly) classes can be unleashed. In real-world application scenarios, the cost of 2D annotations is far lower than that of 3D annotations. Even existing open-world detectors can provide very accurate automatic annotations. To simulate this strategy, we designed the experiments for Table~\ref{2d labels exp kitti} on the KTTTI validation subset. We selected \textbf{\textit{Cars}} and \textbf{\textit{Cyclists}} as known normal categories, and \textbf{\textit{Pedestrians}} as unknown anomalous classes for testing. The 3D boxes for \textbf{\textit{Pedestrians}} are not used during training. Both the 2D annotations from Ground Truth (GT) and the predicted results from the open-vocabulary detector\cite{Groundingdino} were used as supplementary 2D annotations. The results show that the performance brought by the pseudo-labels generated by the existing open-world detector is close to that brought by the ground truth, and due to the open-world detector annotated more objects, its detection performance is even slightly higher than that using the ground truth.

\begin{table}[!ht]
\centering
\caption{
The impact of different 2D box annotation methods on the 3D detection performance of anomaly class Pedestrians.}
\label{2d labels exp kitti}
\begin{tabular}{@{}c|ccc@{}}
\toprule
2D Labels of Ped & $AP3D_{OoD}$ & $AP3D_{car}$ & $AP3D_{cyc}$ \\ \midrule
None             & 0.70          & 80.75        & \textbf{27.63}        \\
KITTI 2D GT      & 20.78         & 80.82        & 22.15        \\ 
Ground DINO\cite{Groundingdino}  & \textbf{21.37}         & \textbf{80.93}        & 23.09        \\\bottomrule
\end{tabular}
\end{table}

\noindent \textit{2) Anomaly Scoring Algorithm.}

\begin{figure*}[!ht]
\centering
\includegraphics[width=7.0in]{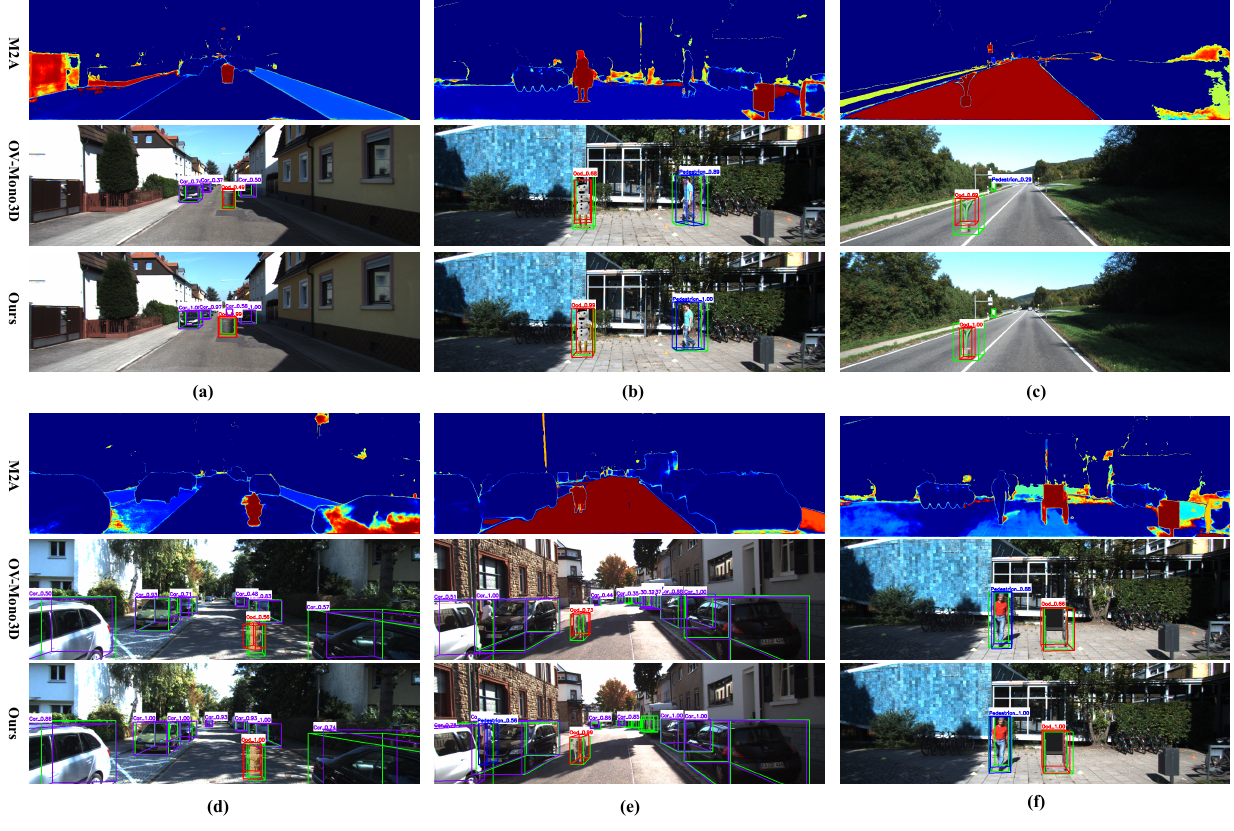}
\hfil
\caption{Comparison of detection results from  Mask2Anomaly, OV-Mono3D and Ours(S3AD), in the 2D frontal-view: (a) trash bin, (b) snowman, (c) flower pot, (d) fire hydrant, (e) sheep, and (f) wheelchair.}
\label{fig_results_2d}
\end{figure*}

The core logic of this paper is to draw inspiration from anomaly segmentation\cite{RBA_ICCV23} and apply it to the anchor level. 
Figure~\ref{fig_results_2d} compares the performance of M2A\cite{Mask2Anomaly_TPAMI24}, OV-Mono3D\cite{OV-Mono3D}, and our proposed S3AD from the 2D front-view perspective. It can be observed that segmentation-based Mask2Anomaly\cite{Mask2Anomaly_TPAMI24} perform pixel-level classification in the 2D space, whereas our method conducts object-level classification in the 3D space. In subfigures (c) and (e) of Figure~\ref{fig_results_2d}, Mask2Anomaly\cite{Mask2Anomaly_TPAMI24} incorrectly segments the ground as an anomalous region, leading to false positives. In contrast, object-level methods like ours are more robust to noise and reduce false alarms. 
Furthermore, 3D approaches offer the additional advantage of estimating collision distances.

\begin{figure}
\centering
\includegraphics[width=3.0in]{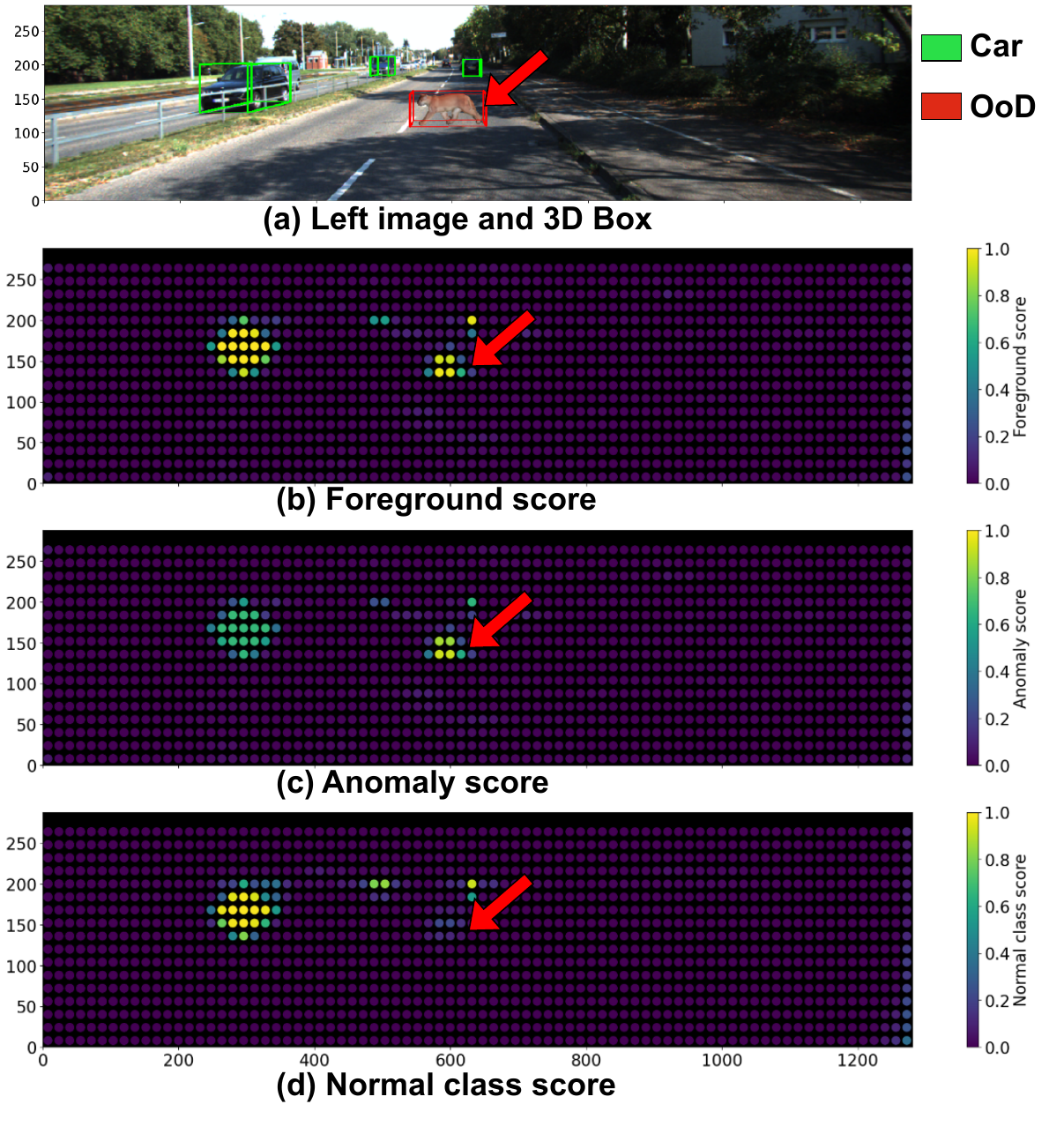}
\caption{Anchor-level Confidence Visualization: (a) left view and detection boxes, (b) foreground confidence score, (c) anomaly class confidence score, (d) normal class confidence score.}
\label{figure-anchors}
\end{figure}

Unlike the dense, pixel-level anomaly scoring used in \textit{Mask2Anomaly}, our method adopts anchor-level anomaly scoring to achieve instance-level anomaly detection. The visualization is shown in Figure~\ref{figure-anchors}. It can be clearly observed that anomalous regions exhibit high foreground and anomaly confidence. 

Table~\ref{Anomaly Score kitti-ar} compares the impact of different anomaly score estimation strategies on detection performance. It can be observed that incorporating foreground confidence improves the overall performance, and the mean-based RbAF strategy achieves the best results.

\begin{table}[ht]
\centering
\caption{Comparison of Different Anomaly Score Estimation Methods  On KITTI-AR-OoD}
\label{Anomaly Score kitti-ar}
\begin{tabular}{@{}c|cccc@{}}
\toprule
Anomaly Score   & $AP_{OoD}$ & $AP_{Ped}$ & $AP_{car}$ & $AP_{cyc}$ \\ \midrule
MSP\cite{MSP}        & 56.96 & 47.83 & 80.18 & \textbf{25.26} \\
MSPF                 & 60.42  & 47.83 & 80.18 & \textbf{25.26} \\
\midrule
RbA\cite{RBA_ICCV23} & 59.14  & 42.78 & 77.83 & 24.97 \\ 
RbAF                 & \textbf{74.35}  & \textbf{48.26} & \textbf{80.20} & \textbf{25.26} \\ \bottomrule
\end{tabular}
\end{table}

\FloatBarrier

\noindent \textit{3) Estimating the Foreground from Feature Disparity.}

\begin{figure*}[!t]
\centering
\includegraphics[width=7.0in]{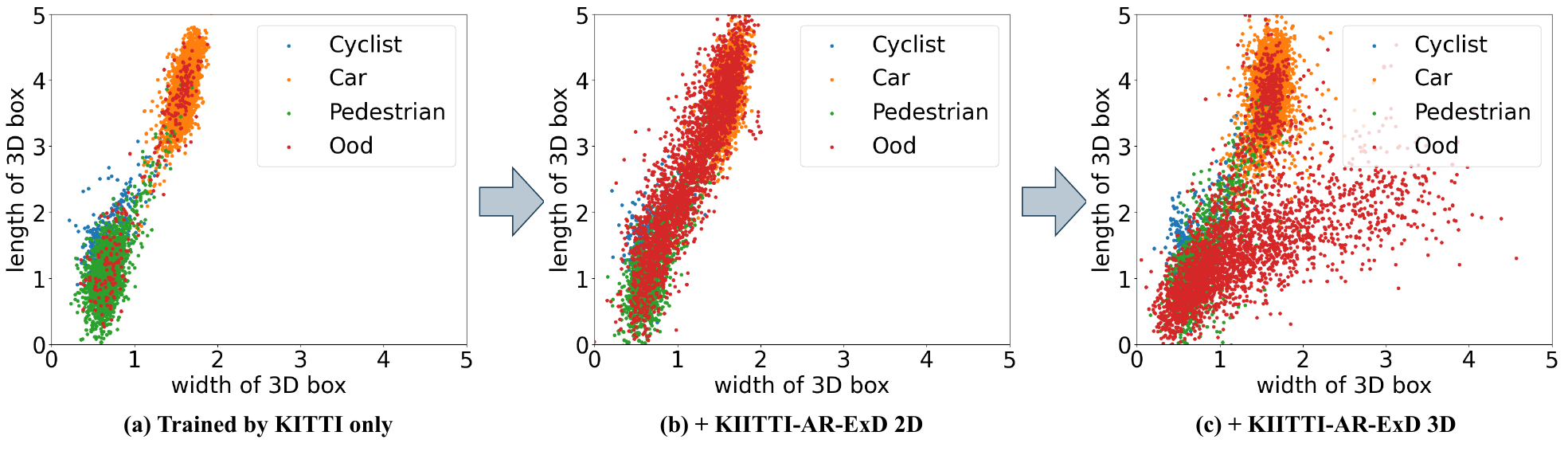}
\hfil
\caption{The scale distribution output by the network on KITTI-AR-OoD under different training datasets: (a) trained on the original KITTI, (b) with the addition of KITTI-AR-ExD 2D Boxes. (c) with the addition of KITTI-AR-ExD 3D Boxes.}
\label{figures output scales}
\end{figure*}

To prevent foreground detection from being overly dependent on the texture of foreground objects in the training set, we propose that 3D foreground detection rely solely on stereo disparity features. This is because foreground objects are closer to the camera and locally protrude compared to the surrounding background. Table~\ref{Fg det input} compares different feature selections for foreground estimation. When foreground estimation is based solely on left stereo features, the 3D detection AP for anomaly categories is 72.73\%. This value improves to 74.35\% when combined with stereo disparity features. When 2D foreground detection is based only on stereo disparity features, its AP still reaches 74.87\%. This indicates that using the disparity features for foreground object detection is more robust.

\begin{table}[ht]
\centering
\caption{Ablation Study on Feature Inputs for Foreground Detection}
\label{Fg det input}
\begin{tabular}{@{}c|cccc@{}}
\toprule
Feature input & $AP3D_{Ood}$ & $AP3D_{Ped}$ & $AP3D_{car}$ & $AP3D_{cyc}$ \\ \midrule
$f_L$         & 72.73   & 46.41  & 79.83  & \textbf{25.76}   \\
$f_s$         & 74.35   & \textbf{48.26}  & \textbf{80.20}  & 25.26   \\
$[f_s, f_L]$  & \textbf{74.87}   & 45.99  & 80.09  & 24.42   \\ \bottomrule
\end{tabular}
\end{table}

\noindent \textit{4) Training 3D Generalization from AR Data.}

To more intuitively demonstrate the impact of the KITTI-AR-ExD dataset on the predictions scale distribution, we have conducted visualizations in Figure~\ref{figures output scales}. 
The figure shows the predicted scale distribution on the KITTI-AR-OoD test set. Red dots indicate the distribution of predicted OoD samples, OoD categories are entirely excluded from training.
Subfigure (a) shows the predicted scale distribution after training only with the original KITTI 2D and 3D boxes. It can be observed that the scale distribution of OoD objects remains similar to that of in-distribution categories from KITTI.
In subfigure (b), additional binary foreground annotations from KITTI-AR-ExD (2D only) are introduced. This leads to improved recall of OoD objects, yet the predicted scales are still overfitted to the known classes.
Subfigure (c) further incorporates 3D box supervision from KITTI-AR-ExD, resulting in predicted scale distributions that better align with the actual object sizes.

\begin{figure}[ht]
\centering
\includegraphics[width=2.4in]{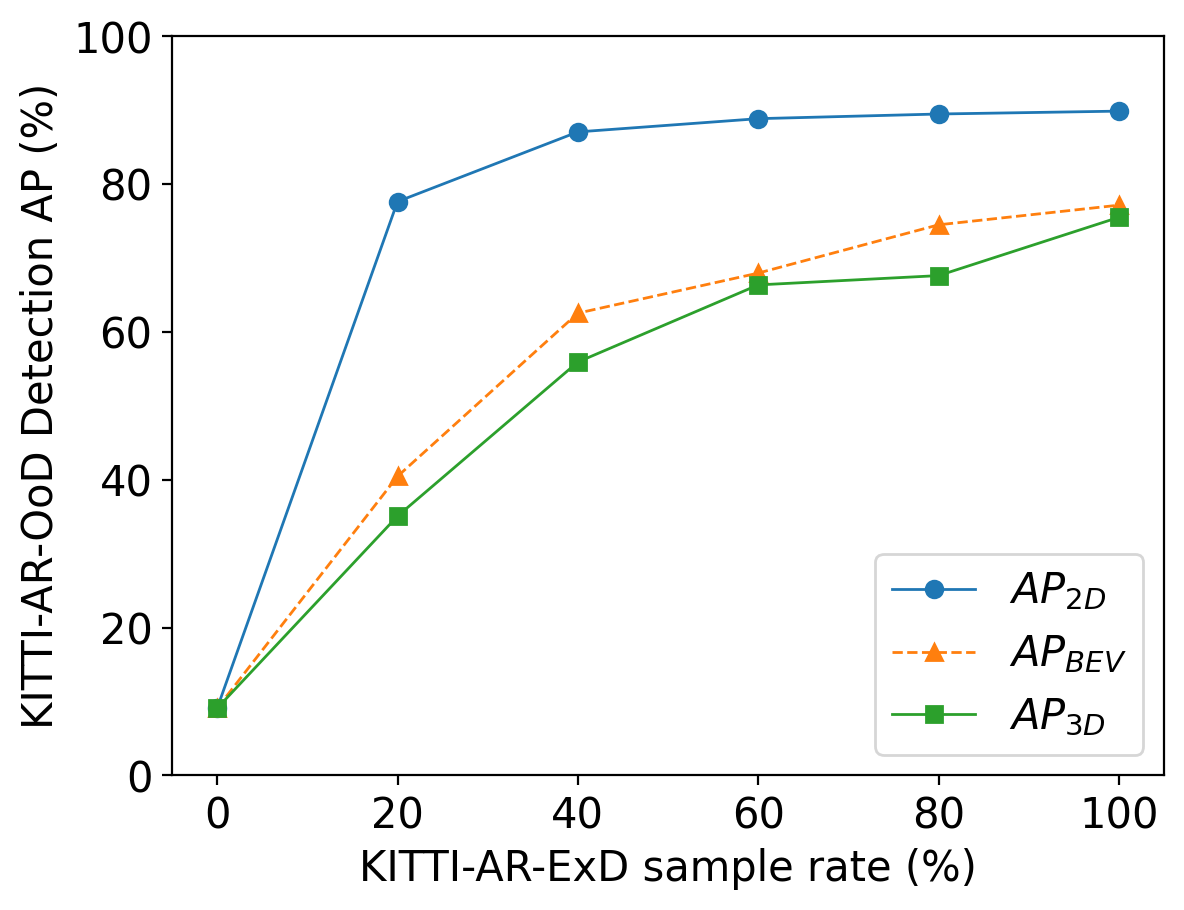}
\caption{The impact of the sampling rate of KITTI-AR-ExD on the AP for OoD(anomaly) detection.}
\label{figures ar nums}
\end{figure}

Figure~\ref{figures ar nums} illustrates the performance trend on OoD novel categories as the AR-assisted training set is progressively expanded. Detailed results on both in-distribution and OoD classes are presented in Table~\ref{kitti_animal_ar_ExD_3D_BEV}. As the number of KITTI-AR-ExD training samples increases, the 2D foreground detection performance improves rapidly and quickly saturates. In contrast, improvements in 3D and BEV detection are more gradual but still significant. Moreover, 3D detection performance on in-distribution categories such as \textbf{\textit{Pedestrian}} and \textbf{\textit{Cyclist}} also improves concurrently, indicating that AR-based synthetic data can serve as an effective form of data augmentation in real-world scenarios, thereby reducing data collection costs.

\begin{table}[ht]
\centering
\caption{The Impact of KITTI-AR-ExD Sample Size on the Test Performance of Detection on KITTI-AR-OoD}
\label{kitti_animal_ar_ExD_3D_BEV}
\begin{tabular}{@{}c|cccc@{}}
\toprule
                   & $AP_{OoD}$* & $AP_{ped}$  & $AP_{car}$  & $AP_{cyc}$  \\
train data         & 3D / BEV    & 3D / BEV    & 3D / BEV    & 3D / BEV    \\ \midrule

 KITTI-train& 9.09/9.09   & 38.48/38.81 & 79.28/79.76 & 20.64/20.64 \\ \midrule
+ExD$\times$0.1 & 24.45/27.02 & 42.88/43.74 & 78.77/79.19 & 21.58/21.58 \\
+ExD$\times$0.2 & 35.08/40.46 & 44.35/44.52 & 79.36/79.82 & 24.25/24.26 \\
+ExD$\times$0.4 & 55.93/62.54 & 46.14/46.45 & 79.34/79.80 & 24.54/24.54 \\
+ExD$\times$0.6 & 66.36/67.96 & 45.78/48.05 & 80.03/80.24 & \textbf{26.20}/\textbf{26.32} \\
+ExD$\times$0.8 & 67.62/74.49 & 47.06/47.84 & \textbf{80.25}/\textbf{80.41} & 24.62/25.00 \\
+ExD$\times$1.0 & \textbf{74.35}/\textbf{76.39} & \textbf{48.26}/\textbf{49.07} & 80.20/\textbf{80.41} & 25.26/25.25 \\ \bottomrule
\end{tabular}
\end{table}

\section{CONCLUSION}
Existing 3D detection algorithms trained on closed-set datasets are incapable of detecting arbitrary road anomalies in an open-world setting. To mitigate the driving risks caused by this phenomenon, this paper improves upon existing stereo 3D detection algorithms by decoupling 2D and 3D detection tasks, unleashing the generalization ability of category-agnostic 3D detection. It also proposes an anomaly scoring strategy based on foreground detection to identify 3D targets of anomalies not seen during training. To validate the effectiveness of the algorithm, two stereo datasets based on augmented reality are proposed. The KITTI-AR-OoD test set fully exposes the algorithm's lack of generalization in scale prediction. The KITTI-AR-ExD training set provides more scale-agnostic category 3D samples during training, significantly enhancing the generalization ability for rare categories and normal categories. Based on this dataset, we will focus on research for faster and more accurate stereo based detection of any 3D objects, such as real-time open-vocabulary 3D detection.

\bibliographystyle{IEEEtran}
\bibliography{ref}

\begin{IEEEbiography}[{\includegraphics[width=1in,height=1.25in,clip,keepaspectratio]{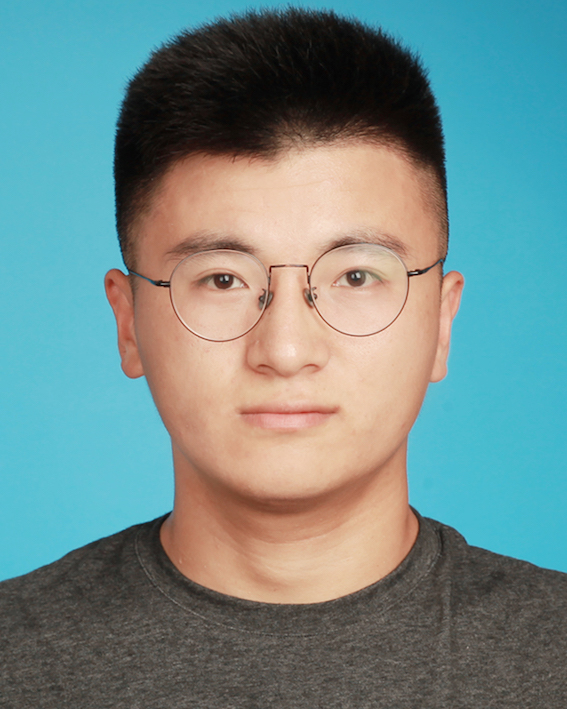}}]{Shiyi Mu}
received the M.Eng. degree from the
School of Communication and Information Engineering, Shanghai University, China, in 2022. He
is currently pursuing the Ph.D. degree with the
information and communication engineering, Shanghai University, China. His research interests include
deep learning for computer vision, optical character recognition, and anomaly detection.
\end{IEEEbiography}
\vspace{11pt}

\begin{IEEEbiography}[{\includegraphics[width=1in,height=1.25in,clip,keepaspectratio]{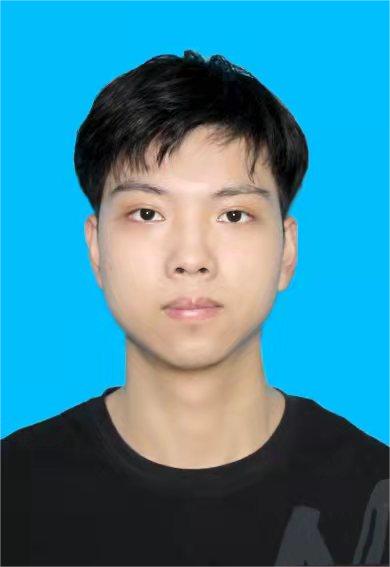}}]{Zichong Gu}
 received the B.Eng. degree from the Department of Communication Engineering, Shanghai University, Shanghai, China, in 2023, where he is currently pursuing the M.Eng. degree with the School of Communication and Information Engineering. His research interests include autonomous driving, depth estimation and open-vocabulary detection.
\end{IEEEbiography}
\vspace{11pt}

\begin{IEEEbiography}[{\includegraphics[width=1in,height=1.25in,clip,keepaspectratio]{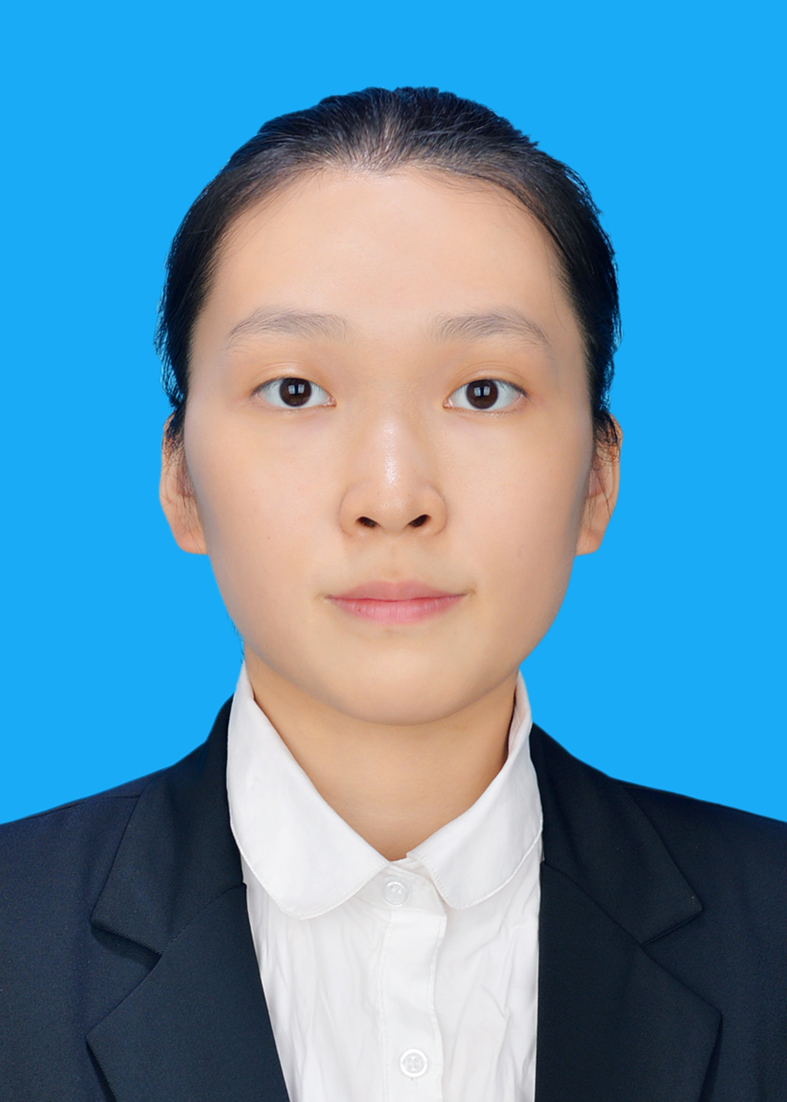}}]{Hanqi Lyu}
is currently pursuing the B.Eng. degree in the Department of Communication Engineering at Shanghai University, Shanghai, China. Her research interests include anomaly detection, depth estimation, and open-vocabulary object detection.
\end{IEEEbiography}
\vspace{11pt}

\begin{IEEEbiography}[{\includegraphics[width=1in,height=1.25in,clip,keepaspectratio]{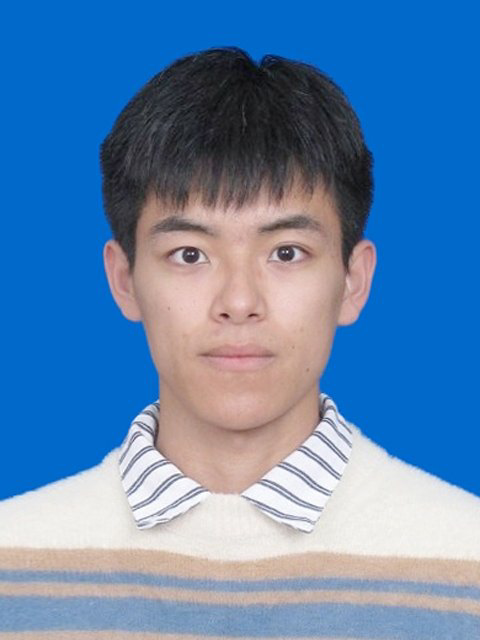}}]{Yilin Gao} received the B.Eng. degree from the Department of Communication Engineering, Shanghai University, Shanghai, China, in 2021. He is currently pursuing the Ph.D. degree in information and communication engineering at Shanghai University, China. His research directions cover OCR, Object Detection, Autonomous Driving, AIGC, and Embodied Intelligence.\end{IEEEbiography}
\vspace{11pt}

\begin{IEEEbiography}[{\includegraphics[width=1in,height=1.25in,clip,keepaspectratio]{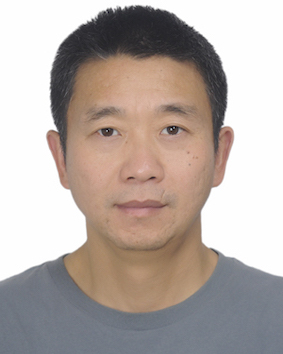}}]{Shugong Xu}
(M'98-SM'06-F'16) graduated from
Wuhan University, China, in 1990, and received his
Master degree in Pattern Recognition and Intelligent
Control from Huazhong University of Science and
Technology (HUST), China, in 1993, and Ph.D.
degree in EE from HUST in 1996. He is now
a professor at Shanghai University. He was the
center Director and Intel Principal Investigator of
the Intel Collaborative Research Institute for Mobile
Networking and Computing (ICRI-MNC), prior to
December 2016 when he joined Shanghai University.
Before joining Intel in September 2013, he was a research director and
principal scientist at the Communication Technologies Laboratory, Huawei
Technologies. He was also the Chief Scientist and PI for the China National
863 project on End-to-End Energy Efficient Networks. Shugong was one of
the co-founders of the Green Touch consortium together with Bell Labs etc,
and he served as the Co-Chair of the Technical Committee for three terms in
this international consortium. Prior to joining Huawei in 2008, he was with
Sharp Laboratories of America as a senior research scientist. Before that, he
conducted research as research fellow in City College of New York, Michigan
State University and Tsinghua University. Dr. Xu published over 160 peer reviewed research papers in top international conferences and journals. He
has over 50 patents granted. He was awarded 'National Innovation Leadership
Talent' by China government in 2013, was elevated to IEEE Fellow in 2015
for contributions to the improvement of wireless networks efficiency. Shugong
is also the winner of the 2017 Award for Advances in Communication from
IEEE Communications Society. His current research interests include machine
learning, pattern recognition, autonomous driving and intelligent machine, as
well as wireless communication systems.
\end{IEEEbiography}

\vfill

\end{document}